\documentclass[letterpaper, 10pt, journal, twoside]{IEEEtran}
\IEEEoverridecommandlockouts
\usepackage{cite}
\usepackage{amsmath,amssymb,amsfonts}
\usepackage{graphicx}
\usepackage{textcomp}
\usepackage{xcolor}
\usepackage{gensymb}
\usepackage{flushend}
\usepackage{xcolor}
\usepackage{icomma}
\usepackage{dblfloatfix} 
\usepackage{algorithm}
\usepackage{algorithmic}
\usepackage{mathtools}
\makeatletter
\let\NAT@parse\relax
\makeatother
\usepackage[colorlinks=true,linkcolor=blue,citecolor=blue,urlcolor=blue]{hyperref}

\title{
Identifying and Exploiting Structure in Robot Co-Design}

\author{Apoorv Vaish$^{1}$ and Oliver Brock$^{1,2,3}$
	\thanks{Manuscript received: May 12, 2026; Accepted: September 1, 2026.}
	\thanks{This paper was recommended for publication by Editor Barbara Mazzolai upon evaluation of the Associate Editor and Reviewers’ comments.} 
	\thanks{$^1$ Robotics and Biology Laboratory, Technische Universit\"at Berlin}%
	\thanks{$^2$ Science of Intelligence (SCIoI), Cluster of Excellence, Berlin}
	\thanks{$^3$ Robotics Institute Germany (RIG)\\
		{\tt\footnotesize apoorv.vaish@tu-berlin.de, oliver.brock@tu-berlin.de}}%
	\thanks{We gratefully acknowledge financial support from the Deutsche Forschungsgemeinschaft (DFG, German Research Foundation) under Germany’s Excellence Strategy – EXC 2002/2 “Science of Intelligence”.
    \newline We used Claude Sonnet 5 for grammar and sentence-structure suggestions.}%
	\thanks{Digital Object Identifier (DOI): see top of this page.}}

\begin{document}
\maketitle

\begin{abstract}

Co-design is a high-dimensional search problem in the robot morphology and control design space. Efficient search requires exploiting the structure shaped by their interaction. To understand this structure, we analyze the landscapes of soft locomotion and manipulation tasks. We identify three patterns consistent across regions of their co-design spaces. Using these insights, we devise an efficient co-design algorithm that yields $36\%$ better co-designs than state-of-the-art baselines. We examine their exploration patterns and show that these baselines required an order of magnitude more function evaluations to find co-designs of comparable quality. Finally, we ablate our algorithm to verify that exploiting the identified structure was the key to efficient co-design.

\end{abstract}


\section{Introduction}
\label{sec:introduction}

\IEEEPARstart{A}{} biological organism exhibits a tight coupling between its body and neural control. This coupling co-adapted over millions of years of evolutionary and developmental processes. In contrast, roboticists typically design morphology and control independently. Changes in one often require changes in the other, making it challenging to maintain a tight coupling. One way to address this challenge is to co-design the robot’s morphology and its control. 

Co-design is a high-dimensional search problem in the combined design space of the robot's morphology and control~\cite{stellaScienceSoftRobot2023, vaishCoDesigningManipulationSystems2024, stolzleSoftEffectiveRobots2025}. Efficient search requires the algorithm to reflect the underlying problem structure~\cite{wolpertNoFreeLunch1997}. However, the structure resulting from the interaction between morphology and control is not well understood, making co-design computationally intensive~\cite{stolzleSoftEffectiveRobots2025, zhangCodesignPowerfulNot2025}.

In this paper, we analyze the optimization landscapes of four co-design tasks (two locomotion and two manipulation). We identify three patterns that hold consistently across regions of their landscapes: 1)~Within a region, quality varies along a low-dimensional manifold, with minimal variation orthogonal to it, reducing the effective search space dimensionality. 2)~In higher-quality regions, the variance in quality is spread across more dimensions, necessitating search to expand dimensionality as quality improves. 3)~In higher-quality regions, quality varies along coupled morphology-control dimensions, requiring search along them.

\begin{figure}[t]
    \centering
    \includegraphics[width=\columnwidth]{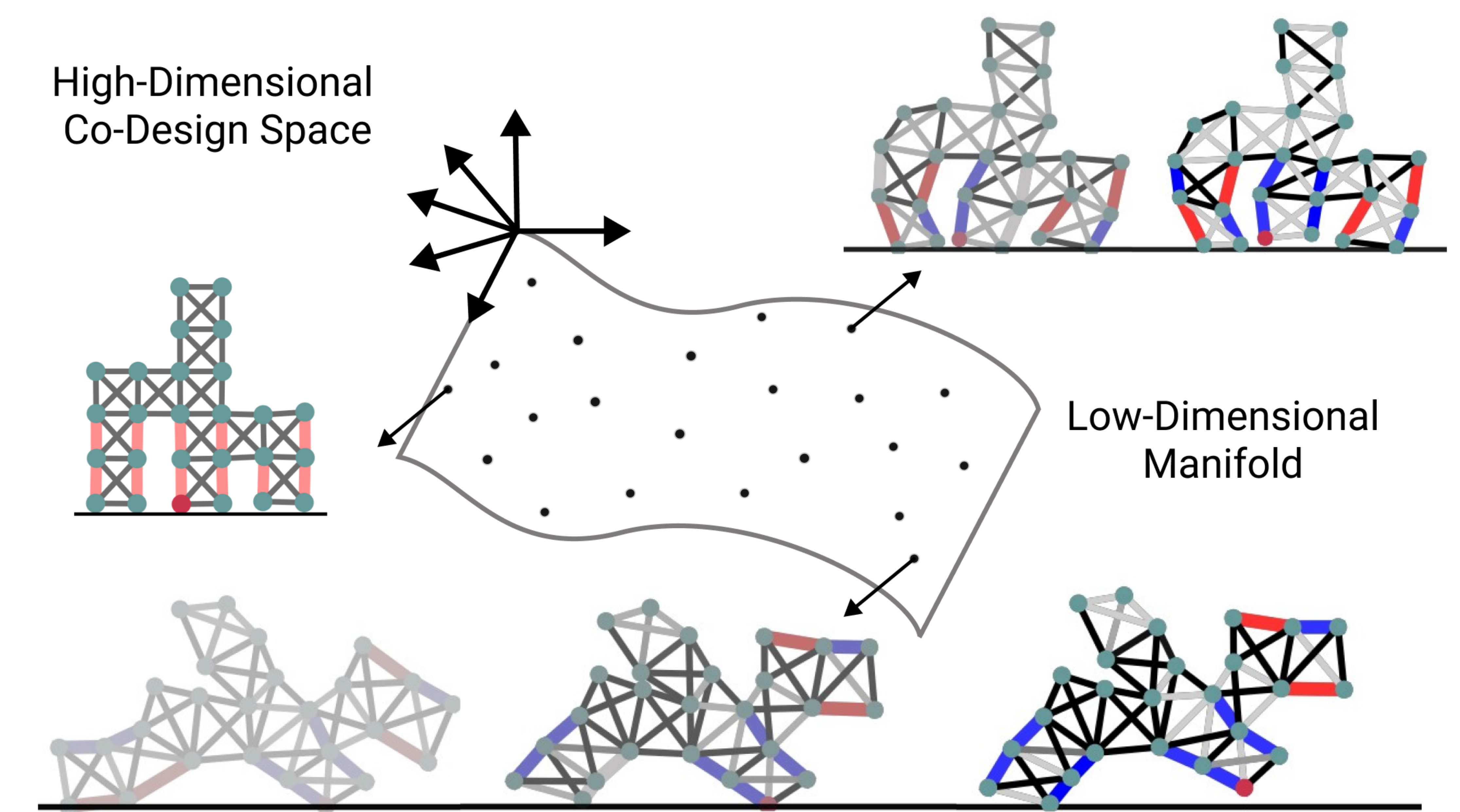}
    \caption{The co-design space, spanned by morphology and control parameters, is high-dimensional. We show that within its regions, quality varies along a low-dimensional manifold. This figure illustrates co-designs found by our algorithm as it exploits this structure during search.}
    \label{fig:teaser}
\end{figure}

We leverage these insights into co-design landscapes to devise an efficient algorithm. This algorithm adaptively identifies and exploits the landscape structure, resulting in a ${36}\%$ improvement in quality over the next-best state-of-the-art baseline across the four tasks. Moreover, in our tests, these baselines required an order of magnitude more function evaluations to yield co-designs of similar quality. In independent evaluation, our algorithm was also recognized as one of the strongest search methods for co-design~\cite{bohlingerShapeYourBody2026}. These results demonstrate the effectiveness of exploiting the identified structure for efficient co-design. 

\section{Related Work}
\label{sec:related-work}

Efficient co-design depends on exploiting the underlying optimization landscape’s structure, which first requires identifying it. Therefore, we compare methods for identifying structure in high-dimensional design spaces and examine what structure existing co-design algorithms exploit.

\subsection{Identifying Structure in High-Dimensional Design Spaces}
\label{sec:related-structure}

High-dimensional design spaces often exhibit an underlying structure that efficient search exploits. Methods for identifying such structure differ in the information they use: the design $x$, the quality $f(x)$, or its gradient $\nabla f(x)$. 

Methods using $x$ identify structure from a dataset of high-quality designs. The most common method, Principal Component Analysis~(PCA), reveals structure in both morphology and control design spaces. For example, it shows that a single variable captures most phenotypic variation in fruit fly wing morphologies~\cite {albaGlobalConstraintsDevelopmental2021}, and that a few synergies largely capture human grasp postures~\cite {santelloPosturalHandSynergies1998}. Nonlinear methods, such as variational autoencoders and graph neural networks, can likewise compactly represent robot morphologies or their motion in low-dimensional latent spaces~\cite{huGLSOGrammarguidedLatent2023, kimLearningRobotStructure2021}. However, the identified structure reflects only the underlying sampling distribution used to generate the high-quality designs, not the optimization landscape. Moreover, collecting such a dataset of high-quality co-designs in itself is computationally intensive~\cite{huGLSOGrammarguidedLatent2023}. Other methods augment $x$ with design quality, $f(x)$, to identify structure. For example, task performance shapes learned representations of robot hand designs~\cite{panEmergentHandMorphology2021}, and soft robot states~\cite{spielbergLearningintheloopOptimizationEndtoend2019}. However, the required number of samples grows exponentially as design parameters increase due to the curse of dimensionality, limiting efficiency in high-dimensional co-design problems.

In contrast, gradients $\nabla f(x)$ directly reveal the landscape's underlying structure. Methods leveraging gradients have been used extensively in engineering design problems, most notably the active subspaces~\cite{lukaczykActiveSubspacesShape2014} method. It computes the covariance of gradients to reveal a subspace along which quality primarily varies. Consequently, the required number of samples grows with the dimensionality of this subspace, and only logarithmically with the design space dimensionality~\cite{constantineComputingActiveSubspaces2015}. This method revealed, for example, that a single dimension dominates the variation in the lift coefficient of a 50-dimensional aerodynamic design space~\cite{lukaczykActiveSubspacesShape2014}. It also revealed insights into the design spaces of solar cells~\cite{constantineDiscoveringActiveSubspace2015b}, lithium-ion batteries~\cite{constantineTimedependentGlobalSensitivity2017}, and thermal design of engines~\cite{greyEnablingAeroengineThermal2019}. However, active subspace methods have not been used to identify the structure resulting from the interaction of morphology and control. Moreover, they assume a global structure, which lets them identify structure from uniformly drawn samples. In contrast, co-design spaces vary regionally, so we first search for regions of varying quality, then use gradients to identify structure within each.

\subsection{Exploiting Structure for Efficient Co-Design}
\label{sec:related-exploiting}

In high-dimensional search problems similar to co-design, reflecting the known structure into the search algorithm has proven effective. For example, the postural synergies discussed above are used to design control strategies for grasping~\cite{ciocarlieHandPostureSubspaces2009} and robot hand morphologies~\cite{catalanoAdaptiveSynergiesDesign2014}. However, the structure of co-design problems is not well understood. As a result, researchers typically rely on standard gradient-based~\cite{xuEndtoEndDifferentiableFramework2021a, vaishCoDesigningManipulationSystems2024} and gradient-free~\cite{deimelAutomatedCodesignSoft2017, luckDataefficientCoAdaptationMorphology2019} algorithms that do not exploit co-design-specific structure.

Black-box optimization algorithms are often augmented with hand-crafted co-design heuristics. For example, co-optimizing morphology and control outperforms optimizing only control~\cite{deimelAutomatedCodesignSoft2017, rosendoTradeoffMorphologyControl2017, vaishCoDesigningManipulationSystems2024}, particularly when the morphology is poorly suited to the task~\cite{zhangCodesignPowerfulNot2025}. Heuristics like morphological innovation protection~\cite{cheneyScalableCooptimizationMorphology2018} temporarily reduce selection pressure on recently modified morphologies, allowing controllers to readapt. Similarly, allocating additional resources to refine the controller after co-optimization further improves performance~\cite{arzaCoOptimizationRobotDesign2024}. However, because the underlying landscapes are not characterized, it remains unclear what structure these heuristics exploit or how to identify new heuristics. Rather than relying on such heuristics, we identify the underlying landscape structure and exploit it for efficient co-design.

\begin{figure}[t]
    \centering
    \includegraphics[width=\columnwidth]{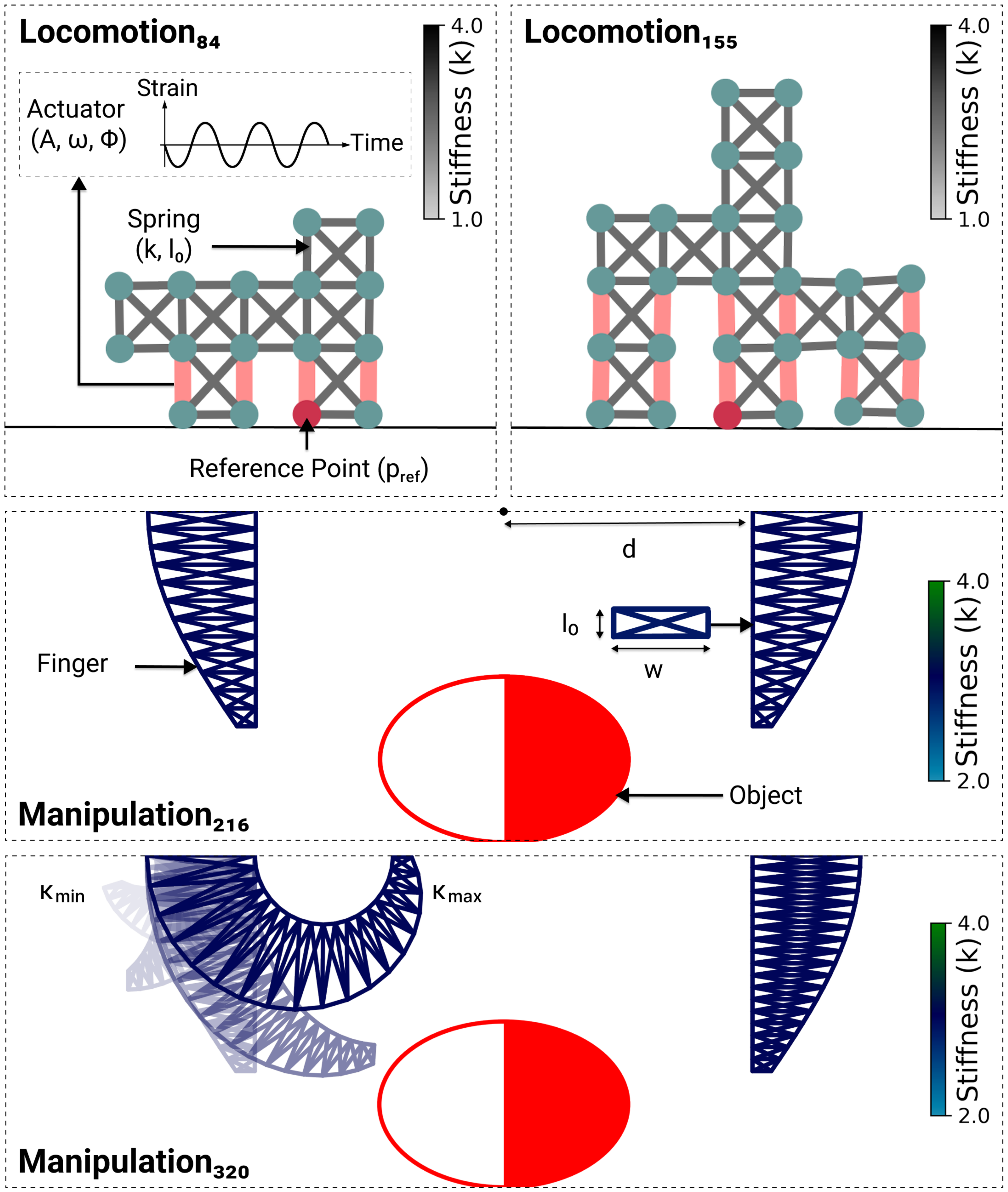}
    \caption{We analyze co-design landscapes for two locomotion tasks (dimensionalities $84$ and $155$) and two manipulation tasks (dimensionalities $216$ and $320$). In locomotion, morphology parameters set each spring's rest length ($l_{0}$) and stiffness ($k$), while control parameters define each actuator's sinusoidal motion ($A, \omega, \phi$).  In manipulation, morphology parameters specify the location of the fingers ($d$), the length ($l_{0}$), and the width ($w$) of each block, and the stiffness ($k$) of each spring, while control parameters set the finger curvature ($\kappa \in [\kappa_{min}, \kappa_{max}]$).}
    \label{fig:problems}
\end{figure}

\begin{figure*}[b]
    \centering
    \includegraphics[width=\textwidth]{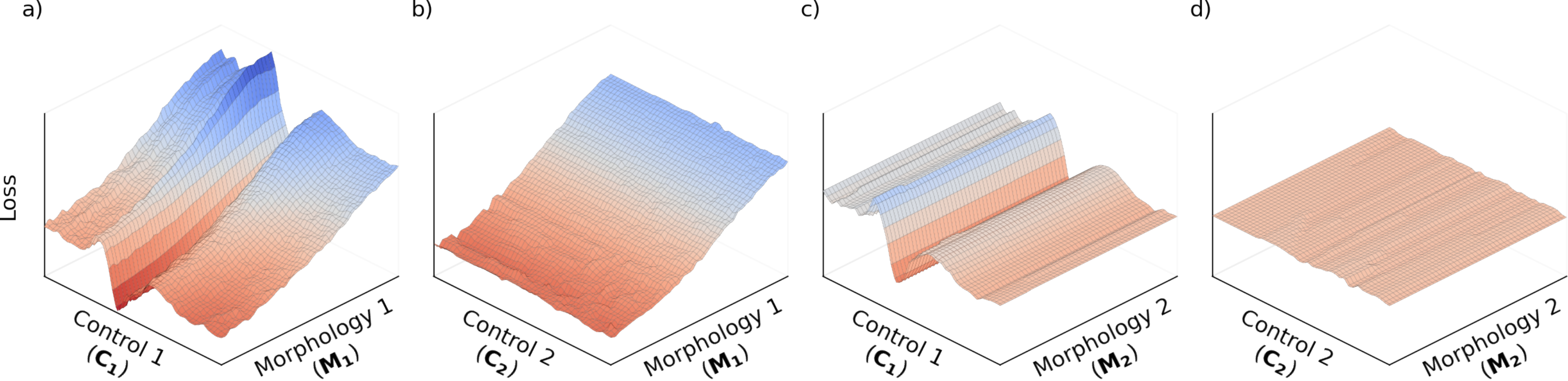}
    \caption{Within a region, quality varies only along a few dimensions, while changing relatively little along others. To illustrate this, we plot landscapes along two-dimensional subspaces spanned by specific morphology~(\textbf{M}) and control~(\textbf{C}) parameters: a) \textbf{M\textsubscript{1}}–\textbf{C\textsubscript{1}}, b) \textbf{M\textsubscript{1}}–\textbf{C\textsubscript{2}}, c) \textbf{M\textsubscript{2}}–\textbf{C\textsubscript{1}}, and d) \textbf{M\textsubscript{2}}–\textbf{C\textsubscript{2}}. We observe that the quality of co-designs primarily varies along \textbf{M\textsubscript{1}} and \textbf{C\textsubscript{1}}, while changing relatively little along \textbf{M\textsubscript{2}} and \textbf{C\textsubscript{2}}.}
    \label{fig:subspace}
\end{figure*}

\section{Co-Design Tasks}
\label{codesign-tasks}

We gain insights into the structure of co-design problems by analyzing their optimization landscapes. Since the landscapes depend on the chosen tasks, we first describe them. To ensure that the identified structure is broadly applicable, our analysis spans both manipulation and locomotion tasks, with two variants of each differing in dimensionality. While these tasks are relatively simple, they capture the complexity of real-world co-design tasks in two ways: 1)~The co-design spaces are sufficiently high-dimensional~(Table~\ref{co-design-problems}) to make exhaustive search infeasible. 2)~Complex interaction between soft morphology and the environment makes manual parameter tuning unintuitive, making computational co-design necessary. We parameterize the co-design spaces with continuous parameters because we rely on gradients for analysis, computed via automatic differentiation in JAX~\cite{bradburyJAXComposableTransformations2018b}. 

\begin{table}[htbp]
    \caption{Co-Design Task Dimensionalities}
    \centering
    \renewcommand{\arraystretch}{1.1}
    \setlength{\tabcolsep}{6pt}
    \begin{tabular}{|c|c|c|c|}
    \hline
    \textbf{Task} & 
    \begin{tabular}[c]{@{}c@{}}\textbf{\# Morphology}   \end{tabular} & 
    \begin{tabular}[c]{@{}c@{}}\textbf{\# Control}      \end{tabular} & 
    \textbf{\# Total} \\
    \hline
    \textit{Locomotion\textsubscript{84}} & 72  & 12 & 84 \\
    \textit{Locomotion\textsubscript{155}} & 122 & 33 & 155 \\  
    \textit{Manipulation\textsubscript{216}} & 200 & 16 & 216 \\
    \textit{Manipulation\textsubscript{320}} & 300 & 20 & 320 \\
    \hline
    \end{tabular}
    \label{co-design-problems}
\end{table}

\subsubsection{Locomotion}
\label{sec:locomotion}

Adapted from DiffTaichi~\cite{huDiffTaichiDifferentiableProgramming2020}, we co-design a soft robot to travel as far as possible. The robot is a two-dimensional lattice of point masses connected by springs, as illustrated in Fig.~\ref{fig:problems}. The two variants, \textit{Locomotion\textsubscript{84}} and \textit{Locomotion\textsubscript{155}}, differ in the number of actuators ($4$ and $11$) and passive springs. We adjust morphology by varying the rest length~$l_{0}$ and stiffness~$k$ of each spring within $\pm20\%$ and $\pm60\%$ of their nominal values. These bounds prevent the lattice from collapsing into unphysical co-designs. A subset of springs are actuators whose $l_{0}$ varies with time, $t$, according to $l_{0}(t) = l_{0}(1 + A\sin(\omega t + \phi))$. $A \in [-0.45, 0.45]$ is the strain amplitude, $\omega \in [-2, 2]$~rad/s is the angular frequency, and $\phi  \in [-\pi, \pi]$~rad is the phase for each actuator.

Forces on the point masses are computed using $F=k(l-l_{0})$, where $l$ is the current spring length. Spring forces are combined with gravity to update velocities of masses using semi-implicit Euler integration with $dt=0.004s$. $\gamma=0.96$ damps velocities, which are integrated to get positions. We resolve collisions with the flat ground using time-of-impact to ensure correct gradients~\cite{huDiffTaichiDifferentiableProgramming2020}. We simulate the robot for $T=2048$ timesteps and measure the final horizontal position ($h$) of a fixed reference point on the robot, $p_{\text{ref}_{h_{T}}}$. The quality function, $f(x)$, is defined as:

\begin{equation}
    f\big(x=
    (\underbrace{k, l_{0}}_{\mathclap{\text{Morphology}}},
    \overbrace{A, \omega, \phi}^{\mathclap{\text{Control}}})\big) = p_{\text{ref}_{h_{T}}}
    \label{eq:locomotion-quality}
\end{equation}

\subsubsection{Manipulation}
\label{sec:manipulation}

We co-design two soft fingers to rotate elliptical objects. Fingers are modeled as linear springs whose stiffness $k$ is adjustable by $\pm33\%$. Six springs forms a block whose length, $l_{0}$, can be adjusted by $\pm25\%$, and width, $w$ by $\pm10\%$. Finally, each finger's position can be adjusted using $d$, bounded to prevent collisions with the object and the environment (the ground and side walls) during initialization. Control parameters set the curvatures, $\kappa_{t} \in [\kappa_{\text{min}}, \kappa_{\text{max}}]$ (shown in Fig.~\ref{fig:problems}), of the upper and lower halves of each finger's spring blocks (two actuators per finger). We update these five times during the first half of the simulation and piecewise-linearly interpolate to obtain the target curvatures, $\kappa_{\text{t}}$, at each timestep. 

 At each timestep, the finger is drawn towards $\kappa_{\text{t}}$. This actuation is combined with gravity to update velocities using semi-implicit Euler integration with $ dt=0.004 \, \mathrm {s} $. Velocities are damped by $\gamma=0.98$ and integrated to get positions. We resolve collisions between the fingers, the object, and the environment using position-based dynamics~\cite{mullerPositionBasedDynamics2007a}. A buckling constraint prevents unphysical self-intersection under large actuation or collision. We simulate the robot for $T=2400$ timesteps, applying random torques to the object after $t=1200$ to test the robustness of the co-designs. The quality function $f(x)$ defined in Eq.~\ref{eq:manipulation-quality}, using the angular error, $\theta_{\text{err}} = \theta_{\text{target}} - \theta_{T}$. Here, $\theta_{\text{target}}$ and $\theta_{T}$ are the target and final object orientations at time $T$ respectively. We choose $\theta_{\text{target}} = 90\degree$ and $ 180\degree$ for the two tasks, \textit{Manipulation\textsubscript{216}} and \textit{Manipulation\textsubscript{320}}, which also differ in the number of springs and therefore in dimensionality. To encourage contact with the object, we add a contact loss, $f_{\text{contact}} = \sum_{\text{fingers}} d_{\text{sdf}}$, where $d_{\text{sdf}}$ is the finger-object signed distance at $t=T$.  We average the sum of both loss components across $n_{\text{envs}}=20$ environments, in which the object is initialized randomly between the fingers, with a $\pm15\%$ change in the length of its major axis to make the tasks challenging.

\begin{equation}
    \label{eq:manipulation-quality}
    f\!\bigl(x=
    (\underbrace{k,l_{0},d,w}_{\mathclap{\text{Morphology}}},
    \overbrace{\kappa_{t}}^{\mathclap{\text{Control}}})\bigr)
    = \tfrac{1}{n_{\text{envs}}}\!\sum_{n_{\text{envs}}}
    \bigl(\theta_{\text{err}}^{2}+f_{\text{contact}}\bigr)
\end{equation}

\begin{figure*}[t]
    \centering
    \includegraphics[width=\textwidth]{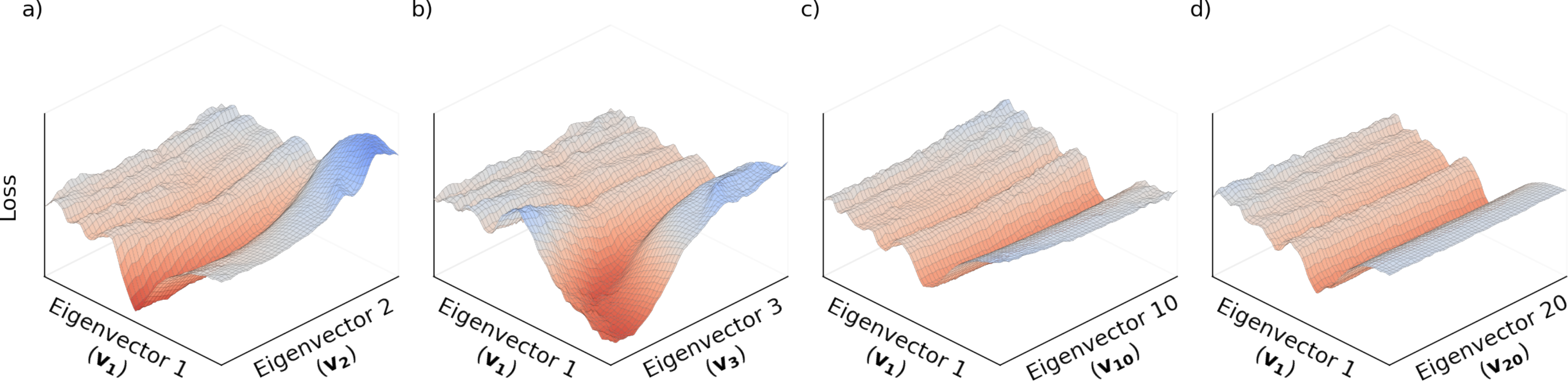}
    \caption{We identify task-relevant dimensions within a region of the co-design space by computing the covariance, $C$, of gradients. The eigenvalues of $C$ rank dimensions by the variance in quality along them. To demonstrate this, we plot two-dimensional subspaces spanned by specific eigenvectors: \textbf{v\textsubscript{1}}–\textbf{v\textsubscript{2}}, \textbf{v\textsubscript{1}}–\textbf{v\textsubscript{3}}, \textbf{v\textsubscript{1}}–\textbf{v\textsubscript{10}}, and \textbf{v\textsubscript{1}}–\textbf{v\textsubscript{20}}. We observe that quality varies the most along \textbf{v\textsubscript{1}}, whereas variation along \textbf{v\textsubscript{20}} is negligible.}
    \label{fig:eigvecs}
\end{figure*}

\begin{figure*}[b]
    \centering
    \includegraphics[width=\textwidth]{img/eigenspectrum.pdf}
    \caption{Within a region of the co-design space, quality varies along a low-dimensional manifold. For each region, we compute the cumulative explained variance in gradients~(see Section~\ref{sec:low-dim}). We plot its mean~(solid line), minimum, and maximum~(shaded area) across regions at each eigenvector index. We observe a steep rise in the eigenspectrum across tasks, indicating that the first few eigenvectors capture most of the variation in gradients.}
    \label{fig:eigenspectrum}
\end{figure*}

\begin{figure*}[t]
    \centering
    \includegraphics[width=\textwidth]{img/dimensionality.pdf}
    \caption{The effective dimensionality~(ED) of sampled gradients increases with the region's quality. For each region, we compute the ED of gradients~(see Section~\ref{sec:dimensionality}) and its quality, measured by the negative loss of the best co-design sampled. We observe a positive linear trend between ED and region quality across all four tasks. This suggests that variance in quality is spread across more dimensions in higher-quality regions.}
    \label{fig:dimensionality}
\end{figure*}

\begin{figure*}[b]
    \centering
    \includegraphics[width=\textwidth]{img/coupling.pdf}
    \caption{The contribution of morphology and control to the gradient becomes more balanced as region quality improves. For each region, we compute its normalized RMS contribution of morphology and control to the gradient~(see Section~\ref{sec:coupling}). We measure the region’s quality by the negative loss of the best co-design sampled. We observe that in low-quality regions (left side of each of the four plots), most contributions come from control parameters (red). As region quality improves (right), the contributions from morphology (blue) and control (red) become more balanced.}
    \label{fig:coupling}
\end{figure*}

\section{Analyzing Co-Design Landscapes}
\label{sec:analyzing-co-design-landscapes}

To exploit the structure of co-design landscapes, we analyze how quality varies across them. Since the space is high-dimensional, exhaustive analysis is intractable. Instead, we look for patterns consistent across its regions.

\subsection{Task-Relevant Dimensions Within a Region}
\label{sec:plotting-landscapes}

To analyze how quality varies within a region of the co-design space, we plot landscapes spanned by two-dimensional subspaces. Fixing all parameters except one morphology (M) and one control (C), we perform a grid search of 2500 points. We pick four such MC subspaces spanned by four co-design parameters ($M_1, M_2, C_1, C_2$) in the task \textit{Locomotion\textsubscript{84}}, and plot the landscapes in Fig.~\ref{fig:subspace}. We observe that quality varies primarily along $M_1$ and $C_1$ with negligible variation along $M_2$ and $C_2$. This indicates that quality varies only along a few task-relevant dimensions within this region.

\subsection{Identifying Task-Relevant Dimensions Using Gradients}
\label{sec:task-relevant-dimensions}

The co-design space is high-dimensional, and the dimensions along which the quality varies might not align with the ambient space. Therefore, we cannot rely solely on landscape plots to identify task-relevant dimensions. To uncover them, we leverage the gradient $\nabla f(x)$ of the quality function $f(x)$ with respect to the co-design parameters $x~\in~\mathbb{R}^n$. While $\nabla f(x)$ quantifies the direction of steepest change in quality at a point, it doesn't capture regional variability. To capture this, we sample gradients at $N$ points, $x_1, \ldots, x_N$, and approximate the uncentered $n \times n$ covariance $C$:
\begin{equation}
    C = \mathbb{E}\!\left[\nabla f(x)\, \nabla f(x)^{\top}\right]
    \approx \frac{1}{N}\sum_{i=1}^{N} \nabla f(x_i)\, \nabla f(x_i)^{\top}
    \label{eq:covariance}
\end{equation}

Since $C$ is symmetric and positive semidefinite, it admits the real eigenvalue decomposition,

\begin{equation}
    C = V \Lambda V^T, \quad \Lambda = \mathrm{diag}(\lambda_1, \ldots, \lambda_m)
    \label{eq:eigendecomp}
\end{equation}

where $V$ contains the eigenvectors and $\Lambda$ the contains the ordered eigenvalues, $\lambda_1 \geq \cdots \geq \lambda_m \geq 0$. To show that $\Lambda$ ranks dimensions by variance, we sample gradients from the region described in Sec.~\ref{sec:plotting-landscapes}. We plot landscapes spanned by the first eigenvector, \textbf{v\textsubscript{1}} and each of \textbf{v\textsubscript{2}}, \textbf{v\textsubscript{3}}, \textbf{v\textsubscript{10}}, and \textbf{v\textsubscript{20}} in Fig.~\ref{fig:eigvecs}. We observe that \textbf{v\textsubscript{1}} captures most of the variation in quality. The remaining eigenvectors capture decreasing variance, with negligible variation along \textbf{v\textsubscript{20}}. This indicates that $\lambda_i$ ranks dimensions by the variance in quality along them, determining their relevance to the co-design task.

\subsection{Sampling Gradients Across Regions of Co-Design Space}
\label{sec:sampling-gradients}

To identify patterns consistent across regions of the co-design space, we sample gradients from $1000$ regions for each task. To maintain diversity in quality and landscape characteristics, we sample regions using trajectories generated by Particle Filter Optimization (PFO, see Sec.~\ref{sec:benchmarks}). We take the best co-design from each iteration as the mean and sample gradients at $N$ points from a Gaussian distribution with $\sigma=0.2$, where parameter values are normalized to $[-1, 1]$. We choose $N = 100, 200, 250, 350$ for each task, ensuring more samples than the co-design space's dimensionality to avoid limiting its rank. We then characterize each region's structure using Eq.~\ref{eq:eigendecomp} to identify patterns consistent across them.

\section{Identifying Landscape Structure}
\label{sec:identifying-structure}

To identify the structure resulting from the interaction between morphology and control, we extract patterns consistent across regions of the co-design space. When leveraged for search, these patterns can help infer information about nearby unexplored regions, guiding search to regions likely to contain promising co-designs.

\subsection{Within Regions, Quality Varies Along a Low-Dimensional Manifold}
\label{sec:low-dim}

We compute the cumulative explained variance in gradients sampled from each region. Fig.~\ref{fig:eigenspectrum} shows the mean, minimum, and maximum cumulative explained variance across regions at each eigenvector index. We observe a steep rise in the eigenspectrum across tasks, indicating that the first few eigenvectors capture most of the variation in gradients. This pattern suggests that within regions of the co-design space, quality varies along a low-dimensional manifold. Since quality varies relatively little orthogonal to the manifold, we can reduce the effective search space. Therefore, within a region, \textbf{search in dimensions along which the quality varies.}

\subsection{Effective Dimensionality Increases with Region Quality}
\label{sec:dimensionality}

We compute the effective dimensionality~(ED) of gradients sampled from a region. It is defined as the Shannon entropy of the eigenspectrum, $\text{ED} = \exp\!\left(-\sum_i p_i \ln p_i\right)$, where $p_i = \lambda_i / \sum_j \lambda_j$ is the normalized eigenvalue~\cite{delgiudiceEffectiveDimensionalityTutorial2021}. We measure a region's quality by the negative loss of the best co-design sampled from the region. We bin regions by their quality and plot the mean and standard deviation of ED within each bin in Fig.~\ref{fig:dimensionality}. We observe that higher-quality regions exhibit greater ED. While there is some variance within a bin, this observation is largely consistent across regions and tasks. This pattern indicates that in higher-quality regions, variation in quality is spread across more dimensions. Therefore, \textbf{search in more dimensions as quality improves.} 

\begin{figure*}[b]
    \centering
    \includegraphics[width=\textwidth]{img/benchmarks.pdf}
    \caption{Identifying and exploiting the structure of co-design landscapes enables GC-PFO to find better co-designs across tasks consistently. We compare our algorithm, Gradient Covariance Particle Filter Optimization (GC-PFO), with conventional gradient-based (SGD, Adam) and gradient-free (PFO, CMA-ES, GA) algorithms. Across these tasks, GC-PFO yields $36\%$ better co-designs than the next-best algorithm.}
    \label{fig:benchmarks}
\end{figure*}

\subsection{Morphology and Control Contributions to the Gradient Become Balanced as Region Quality Improves}
\label{sec:coupling}

We quantify the contributions of morphology and control to gradients sampled from a region. We compute the root-mean-square (RMS) magnitudes of morphology and control components to account for differences in their dimensionalities. We average these across all gradients sampled from a region. We normalize the resulting RMS magnitudes so that their sum equals $1.0$, yielding their contribution to the gradient. We measure a region's quality by the negative loss of the best co-design sampled from the region. We bin regions by quality and plot the mean and standard deviation of contributions within each bin in Fig.~\ref{fig:coupling}. As region quality improves, morphology and control contributions become more balanced. While there is some variation across regions of similar quality, this pattern is largely consistent across all four tasks. This pattern indicates that in higher-quality regions, quality varies comparably along both morphology and control parameters, requiring \textbf{search along coupled morphology-control dimensions}.

\section{Exploiting Structure for Co-Design}
\label{sec:exploiting-structure}

Since search performance depends on reflecting the problem structure in the algorithm~\cite{wolpertNoFreeLunch1997}, efficient co-design depends on exploiting the structure identified in Sec.~\ref{sec:identifying-structure}. However, we must augment this with task-specific structure, specifically the dimensions along which quality varies. These dimensions are not known a priori, but can be inferred from gradients gathered during search. Sec.~\ref{sec:task-relevant-dimensions} shows that the gradient covariance matrix $C$ reveals them. We incorporate this task-specific structure into PFO~\cite{liuParticleFilterOptimization2016b} to devise Gradient Covariance Particle Filter Optimization (GC-PFO).  GC-PFO combines gradient-based structure identification with gradient-free PFO: the former identifies low-dimensional structure, while the latter helps avoid local minima. PFO is well suited for this, as information about the region's structure can bias a particle's exploration toward dimensions where quality varies most. Specifically, we use $\beta$, 
 
\begin{equation}
    \label{eq:bias}
    \beta = V \tilde{\Lambda}, \text{ where }
    \tilde{\Lambda}
    = \frac{\Lambda}{\frac{1}{n}\operatorname{tr}(\Lambda) + \epsilon}
\end{equation}

Here, $\tilde{\Lambda}$ normalizes the eigenvalues of $C$ by their mean. $\beta$ combines these with the eigenvectors $V$ to bias particle exploration toward dimensions along which quality varies most. $\epsilon$ prevents division by zero. The algorithm maintains $N$ particles distributed across $R$ regions. At each iteration, it evaluates each particle's quality and gradient, then resamples particles based on their quality. Particles are then updated using $\beta$. We summarize these steps in Algorithm~\ref{alg:gc-pfo}.

\subsection{Benchmarks}
\label{sec:benchmarks}

We evaluate GC-PFO on the four tasks described in Section~\ref{codesign-tasks}. We compare its performance against two gradient-based optimizers: SGD with momentum and Adam~\cite{kingmaAdamMethodStochastic2017}, implemented using Optax, and three gradient-free algorithms, PFO~\cite{liuParticleFilterOptimization2016b}, CMA-ES~\cite{hansenAdaptingArbitraryNormal1996}, and Genetic Algorithm (GA)~\cite{hollandGeneticAlgorithmsOptimal1973}, implemented using evosax. For gradient-based methods, we initialize each restart by uniformly sampling the co-design space with $\sigma=0.5$. For the other algorithms, we initialize the distribution means similarly. Each algorithm evaluates the quality function $250$ times for \textit{Locomotion\textsubscript{84}}, and $500$ times for the other tasks per iteration. To find the best learning rates for SGD and Adam, we tested $10$ values across $3$ orders of magnitude. For CMA-ES and GA, we tested $\sigma=0.1, 0.2, 0.4, 0.8$. For each $\sigma$, we tested two population sizes: $50$ and $250$ for \textit{Locomotion\textsubscript{84}}, and $50$ and $500$ for other tasks, ensuring an equal number of function evaluations and using the default elite ratio. For PFO and GC-PFO, we divided particles into regions containing $50$ particles each and set $\sigma=0.2$. We ran each algorithm $5$ times with different seeds and report the mean best loss across these runs in Fig.~\ref{fig:benchmarks}.

\begin{algorithm}[H]
    \caption{Gradient Covariance Particle Filter Optimization~(GC-PFO)}
    \label{alg:gc-pfo}
    \begin{algorithmic}
    \STATE \textbf{Input:} Number of regions ($R$), particles per region ($P$), step size ($\sigma$), annealing rate ($\kappa$), iterations ($iters$)
    \STATE \textbf{Initialize} $R$ regions with $P$ particles around mean $x_{r,0}$
    \FOR{$i = 0$ to $iters - 1$}
        \FOR{each region $r$}
            \STATE \textbf{Evaluate:} Compute losses $L_{r,j}$ and gradients $\nabla f(x_{r,j})$ for all particles $x_{r,j}$
            \STATE \textbf{Resample:} Sample particles with probability $p_{r,j}\propto \exp(-\frac{L_{r,j}}{\tau_t})$, with temperature, $\tau_t = \frac{1}{1 + \kappa \cdot \left(\frac{i}{iters}\right)}$
            \STATE \textbf{Identify Task-Relevant Dimensions:}
            \STATE \quad Compute $C_r \approx G_r G_r^{\top}$ using Eq.~\ref{eq:covariance}
            \STATE \quad Decompose $C_r = V_r \Lambda_r V_r^{\top}$ using Eq.~\ref{eq:eigendecomp}
            \STATE \quad Compute $\beta_r = V_r \tilde{\Lambda}_r$ using Eq.~\ref{eq:bias}
            \STATE \textbf{Update Particles:} For each particle, draw Gaussian noise, $\eta_{r,j}\sim\mathcal{N}(0,I)$, and set $x_{r,j}\gets x_{r,j} +  \sigma\,\beta_r\,\eta_{r,j}$
        \ENDFOR
    \ENDFOR
    \STATE \textbf{Return} best particle $x^*$
    \end{algorithmic}
\end{algorithm}

\begin{figure*}[b]
    \centering
    \includegraphics[width=\textwidth]{img/exploration.pdf}
    \caption{GC-PFO discovers high-quality regions that benchmark algorithms do not reach efficiently~(Sec.~\ref{sec:exploration}). \textbf{a)} Co-Design Space Exploration: t-SNE visualization of the explored co-design space for \textit{Locomotion\textsubscript{155}} shows that GC-PFO explores high-quality regions that are unexplored by the benchmark algorithms. \textbf{b)} Sample Efficiency: Benchmark algorithms require more than an order of magnitude more function evaluations to find co-designs of comparable quality to GC-PFO. \textbf{c)} Ablation~(Sec.~\ref{sec:ablation}): discarding the low-dimensional structure~(PFO) and shrinking the search dimensionality~(GC-PFO~w/o~P2) degrades performance the most, while restricting search to morphology or control dimensions~(GC-PFO~w/o~P3) has the least impact.}
    \label{fig:exploration}
\end{figure*}

\subsection{Results}
\label{sec:results}

\begin{figure}[t]
    \centering
    \includegraphics[width=\columnwidth]{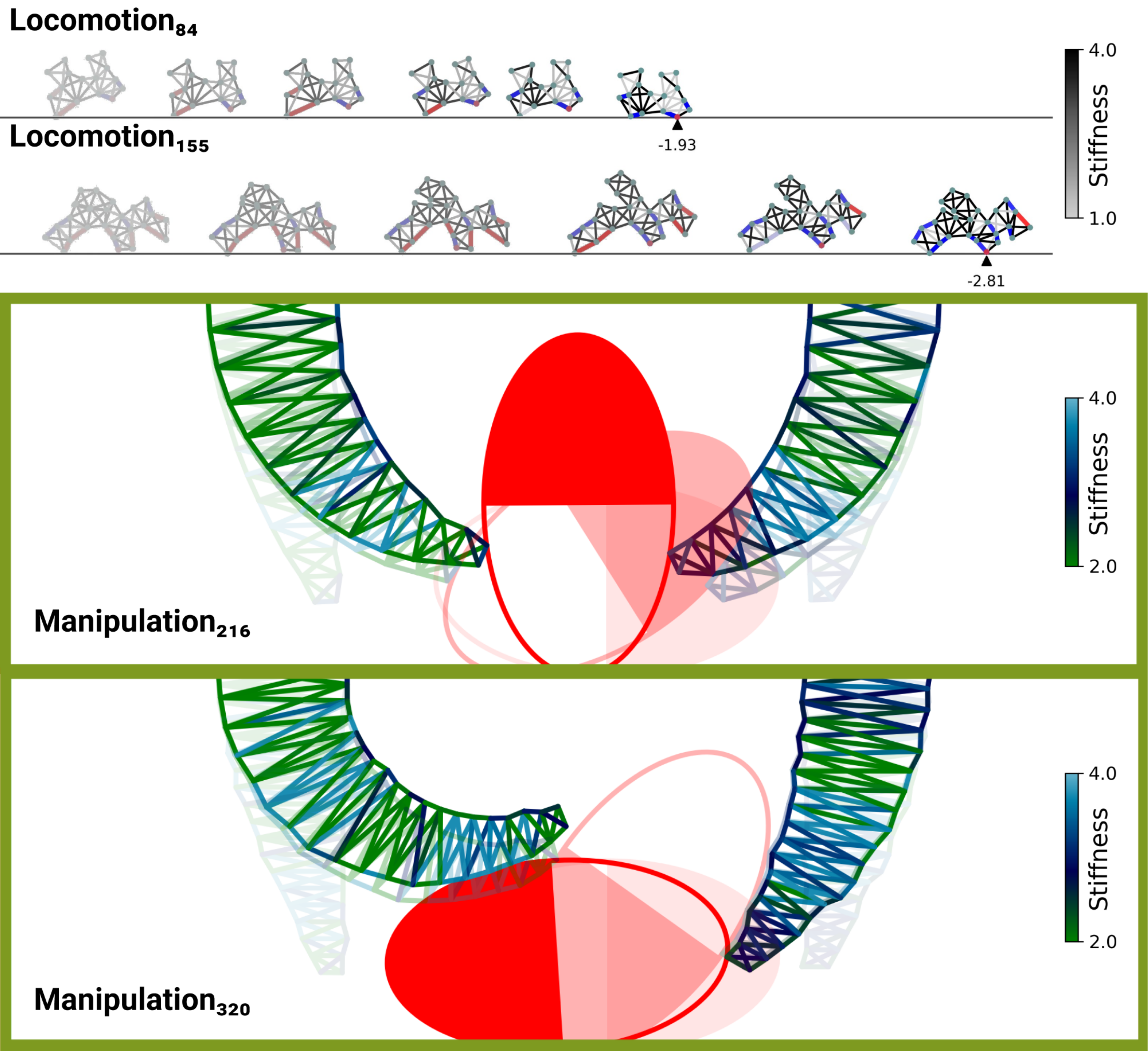}
    \caption{Best co-designs found by our algorithm, Gradient Covariance Particle Filter Optimization~(GC-PFO). In \textit{Locomotion\textsubscript{84}} and \textit{Locomotion\textsubscript{155}}, GC-PFO finds co-designs in which morphologies complement effective leg synchronization. In \textit{Manipulation\textsubscript{216}}, the robot first closes the gap between the fingers to reduce uncertainty in the initial object pose and size. One finger is slightly smaller, helping pivot and rotate the object. In \textit{Manipulation\textsubscript{320}}, GC-PFO finds a relatively compliant (left) fingertip that pivots the object, while the stiff finger catches it. The fingers then constrain it against the ground, resisting external torque.}
    \label{fig:results}
\end{figure}

We measure each algorithm's performance as the percentage improvement in loss from the initial value. GC-PFO outperforms the benchmark algorithms across all four tasks. In \textit{Locomotion\textsubscript{84}} and \textit{Locomotion\textsubscript{155}}, GC-PFO achieves around $29\%$ and $54\%$ better co-designs over the next-best algorithm, GA and CMA-ES, respectively. It produces $25\%$ and $39\%$ better co-designs in \textit{Manipulation\textsubscript{216}} and \textit{Manipulation\textsubscript{320}}. SGD and Adam plateau after a few iterations across all tasks, likely because gradient-based methods struggle with local minima as the manifold's effective dimensionality increases. We show the best co-designs found by GC-PFO in Fig.~\ref{fig:results}. GC-PFO was also independently identified as one of the strongest search methods for co-designing practical robots, including Unitree Go2, MIT Humanoid, and ANYmal C~\cite{bohlingerShapeYourBody2026}. These results demonstrate the effectiveness of exploiting the identified structure for co-design.

\subsection{Co-Design Space Exploration}
\label{sec:exploration}

To understand why benchmark algorithms cannot find co-designs of comparable quality to GC-PFO, we consider two hypotheses: either they fail to find high-quality regions, or they get stuck in local minima after finding them. To test this, we examine their exploration patterns in \textit{Locomotion\textsubscript{155}} using t-SNE. Fig.~\ref{fig:exploration}~a. reveals that GC-PFO discovers high-quality regions that other algorithms do not. Because they do not explore these regions within the evaluation budget, they can reach them only through random restarts, which require additional function evaluations.

We quantify the additional function evaluations the benchmark algorithms need to close the performance gap. We set a budget of $50,000$ evaluations for GC-PFO, whereas the benchmark algorithms can perform $1000$ times as many evaluations using the best hyperparameters from Sec.~\ref{sec:benchmarks}. To avoid wasting computational effort on unproductive restarts, each restart terminates when the best loss hasn’t changed for the last $10$ iterations or when $250$ iterations have elapsed. We plot the best loss found against the number of function evaluations in Fig.~\ref{fig:exploration}~b. We observe that CMA-ES approaches GC-PFO after approximately 80 times as many function evaluations. GA and Adam find co-designs of the same quality as GC-PFO in $270$ and $950$ times as many function evaluations, respectively. These results demonstrate GC-PFO's effectiveness in finding high-quality regions.

\subsection{Ablation}
\label{sec:ablation}

To verify that exploiting the identified structure is the key to efficient co-design, we design three GC-PFO ablations, each removing one structural pattern identified in Sec.~\ref{sec:identifying-structure}. Removing the first pattern collapses to an isotropic search (i.e., PFO). Removing the second pattern yields GC-PFO~w/o~P2 that shrinks the rank of $C$ by zeroing out eigenvalues beyond $k$. $k$ is reduced linearly from the ambient dimensionality to one over iterations, as a proxy for reducing dimensionality as quality improves. Removing the third pattern yields GC-PFO w/o P3, which randomly restricts half the particles' updates to morphology dimensions and the other half to control dimensions during each iteration, enabled only after the first $20$ iterations to target high-quality regions. We run these variants on Locomotion\textsubscript{155} with a budget of \(50,000\) function evaluations. We plot the best loss found so far, averaged across $5$ seeds, against the number of function evaluations in Fig.~\ref{fig:exploration}~c. 

We observe that PFO yields \(58\%\) worse improvements than GC-PFO from the initial quality, indicating that leveraging the low-dimensional structure improves performance. GC-PFO~w/o~P2 trails GC-PFO by \(50.1\%\), providing evidence consistent with the second pattern. GC-PFO~w/o~P3 trails GC-PFO by \(14.9\%\), indicating that searching in coupled dimensions has a modest impact. This is likely because the particles can still exploit the low-dimensional structure independently in morphology and control subspaces over successive iterations. Overall, these results are consistent with the identified patterns, demonstrating their effectiveness in guiding search.

\section{Conclusions}
\label{sec:conclusions}

We analyzed co-design landscapes and showed they exhibit exploitable structure. It remains to be seen whether similar structure emerges with alternate representations. Our analysis also revealed that quality varies along coupled morphology-control dimensions, indicating that they interact strongly to affect performance. This suggests that novel representations that capture their synergistic interactions might be better suited for co-design. Finally, we demonstrated the effectiveness of identifying and exploiting landscape structure for efficient co-design. This could be augmented with task-specific knowledge from prior data or expert knowledge, allowing the search to be seeded in high-quality regions, making practical co-design problems tractable.

\makeatletter
\@ifundefined{l@english}{\chardef\l@english=0\relax}{}
\let\l@en\l@english
\let\l@eng\l@english
\makeatother
\bibliographystyle{IEEEtran}
\bibliography{RAL26}

\end{document}